\documentclass[letterpaper, 10 pt, conference]{ieeeconf} 

\IEEEoverridecommandlockouts 
\usepackage{amsmath}
\usepackage{algorithm}
\usepackage{algorithmic}
\usepackage{caption}
\usepackage{graphicx, subfig}
\usepackage{multirow}
\usepackage[table,xcdraw]{xcolor}
\usepackage{colortbl}
\usepackage{float} 
\renewcommand{\baselinestretch}{0.97} 

\title{Learning to Perform Low-Contact Autonomous Nasotracheal Intubation by Recurrent Action-Confidence Chunking with Transformer}

\author{Yu Tian$ ^{1\dag}$, Ruoyi Hao$ ^{1\dag}$, Yiming Huang$ ^{1}$, Dihong Xie$ ^{1}$,\\ Catherine Po Ling Chan$^{2}$, Jason Ying Kuen Chan$^{2}$, Hongliang Ren$ ^{1*}$ 
\thanks{This work is supported by the Hong Kong Research Grants Council (RGC) Collaborative Research Fund (CRF-C4026-21G) and the  Regional Joint Fund Project of the Basic and Applied Research Fund of Guangdong Province (2021B1515120035).}
\thanks{We would like to thank Hang Li, Yupeng Wang, Jianwei Zheng, Yongze Wu, Qingyang Liu, and Yang Yang for participating in the experiment.}
\thanks{$ ^{\dag}$ These authors contribute equally to this work.}
\thanks{$ ^{1}$ The authors are with the Department of Electronic Engineering, The Chinese University of Hong Kong, Shatin, N.T., Hong Kong SAR, China. }
\thanks{$ ^{2}$ The authors are with the Department of Otorhinolaryngology, Head and Neck Surgery,
The Chinese University of Hong Kong, Shatin, N.T., Hong Kong SAR, China. }
\thanks{*Corresponding author: {\tt\small hlren@ee.cuhk.edu.hk}}}

\begin{document}
\maketitle
\thispagestyle{empty}
\pagestyle{empty}

\begin{abstract}
Nasotracheal intubation (NTI) is critical for establishing artificial airways in clinical anesthesia and critical care. Current manual methods face significant challenges, including cross-infection, especially during respiratory infection care, and insufficient control of endoluminal contact forces, increasing the risk of mucosal injuries. While existing studies have focused on automated endoscopic insertion, the automation of NTI remains unexplored despite its unique challenges: Nasotracheal tubes exhibit greater diameter and rigidity than standard endoscopes, substantially increasing insertion complexity and patient risks. We propose a novel autonomous NTI system with two key components to address these challenges. First, an autonomous NTI system is developed, incorporating a prosthesis embedded with force sensors, allowing
for safety assessment and data filtering. Then, the Recurrent Action-Confidence Chunking with Transformer (RACCT) model is developed to handle complex tube-tissue interactions and partial visual observations. Experimental results demonstrate that the RACCT model outperforms the ACT model in all aspects and achieves a 66\% reduction in average peak insertion force compared to manual operations while maintaining equivalent success rates. This validates the system's potential for reducing infection risks and improving procedural safety.
\end{abstract}

\section{Introduction}

\subsection{Nasotracheal Intubation}

Intubation is a critical medical procedure used in various clinical settings, particularly in emergency medicine, critical care, and anesthesiology  \cite{torrego2020bronchoscopy, tobin2020basing, deng2024assisted}. It involves inserting a tube through the patient's airway to maintain an open-air passage, facilitate breathing, or deliver anesthetic gases  \cite{murakami2023therapeutic}. The procedure is essential for patients requiring mechanical ventilation due to respiratory failure  \cite{licker2007perioperative}, during surgeries under general anesthesia, or in emergencies where airway protection is crucial. There are two main types of intubation: orotracheal and nasotracheal. Nasotracheal intubation is often preferred in situations such as oral surgeries, dental procedures, and cases requiring prolonged intubation, as it offers better surgical access, enhanced conscious patient comfort and tolerance, and is more suitable for long-term use and nursing care compared to orotracheal intubation  \cite{christian2020use}. 

NTI’s technical complexity, stemming from rigid tubes and narrow nasal anatomy \cite{chauhan2016nasal}, increases mucosal injury risks \cite{nan2023characteristics} (e.g., epistaxis, sinus damage) and elevates the risk of procedural errors. Improper tube placement or prolonged attempts may further provoke laryngospasm, hypoxia, or, in severe cases, brain damage \cite{sun2010cardiovascular}.
Moreover, during intubation, medical staff are directly exposed to possible viruses and contaminated aerosols from patients  \cite{gasmi2022improving}, a risk highlighted by the recent COVID-19 pandemic  \cite{izzetti2020covid}.

Improving intubation techniques can enhance patient safety, outcomes, and healthcare worker protection. This has led to the exploration of robotic technology  \cite{dupont2021decade, ng2024navigation} to assist with intubation procedures. 
While various robotic systems have been developed for orotracheal intubation  \cite{hemmerling2012first,wang2018original,liang2020pneumatic, ponraj2022chip, lai2023sim, wang2023domain}, robot-assisted nasotracheal intubation systems are still in their early stages. 
Existing NTI robots \cite{deng2023safety, deng2023automatic, hao2025variable} focus on endoscope-guided lumen tracking with visual collision avoidance during the initial navigation step. While our previous work \cite{hao2025variable} introduced passive safety mechanisms to mitigate collisions at this step, mucosal injuries predominantly occur in the subsequent NTT insertion step. This is due to the tube’s larger diameter, higher rigidity compared to endoscopes, and the lack of self-feedback, highlighting a critical safety gap in current robotic NTI designs.

\begin{figure*}[t]
  \centering
  \includegraphics[width=1\linewidth]{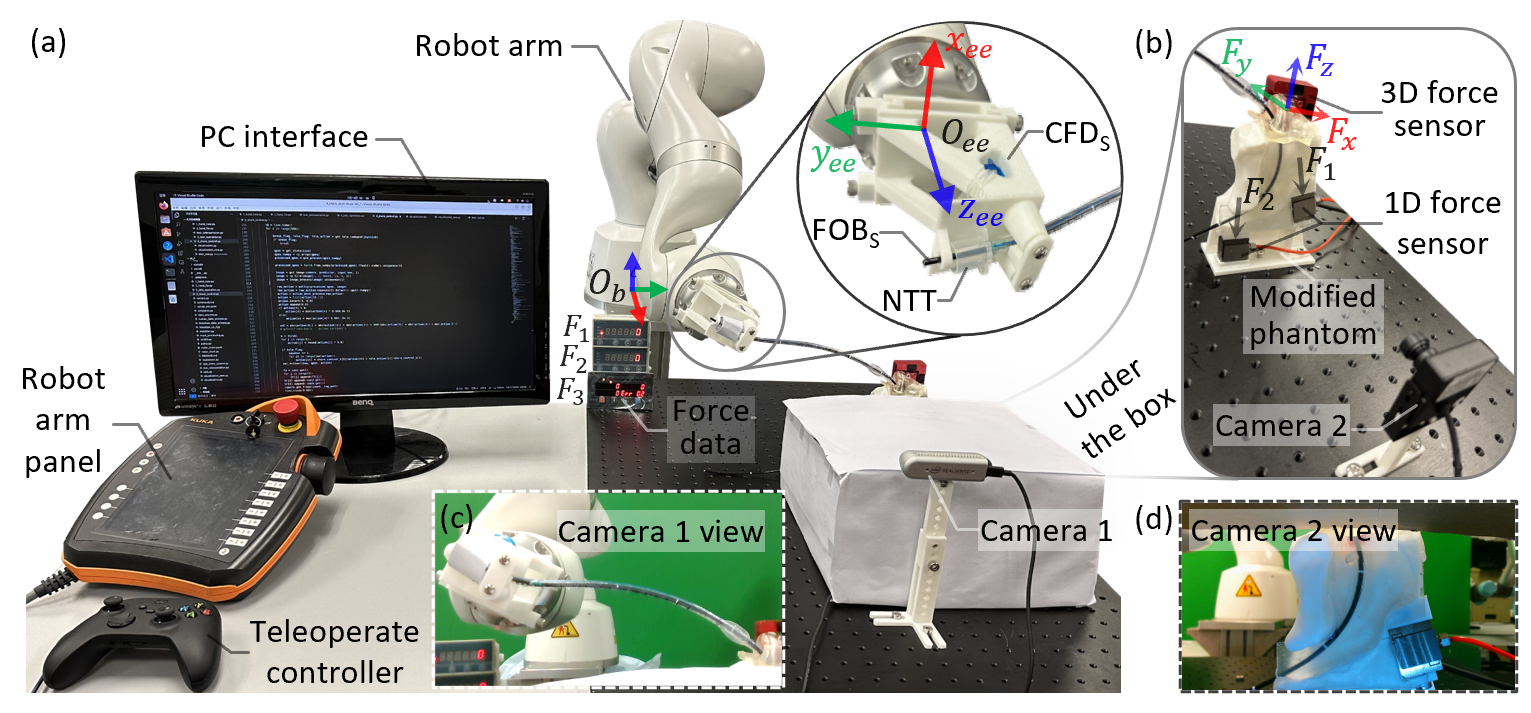}
  \caption{\textbf{Hardware overview.} The hardware system consists of a KUKA Iiwa robot and its panel, 2 cameras, a nasotracheal tube, a PC, a teleoperation controller, a prosthesis with three force sensors embedded in it, and 3 force monitors.}
  \label{system}
\end{figure*}

\subsection{Segmentation-assisted Tube Tracking.}
Segmenting the tracheal tube outside the oropharyngeal region is crucial for accurate tracking and automation of intubation procedures. The tracheal tube is a relatively thin, cylindrical, deformable linear object (DLO). Advanced image processing techniques such as edge detection, segmentation, and shape analysis are essential to track the deformation of this DLO during the procedure effectively. These techniques help identify the tube’s location and distinguish it from the surrounding background, the surgical robot arm, and other elements. 

To address the issue of complex environments, previous works \cite{zanella2021auto, thananjeyan2022all, caporali2022fastdlo} introduced learning-based DLO segmentation, achieving a more robust DLO segmentation in varying backgrounds. However, these methods are still not robust against the non-uniform color situation caused by the transparent layer of the tracheal tube. Collecting large-scale training data with the transparent tube is a feasible solution to this challenge. However, the data collection and labeling or rendering in simulation software takes human effort and may perform poorly in unseen surgical scenarios. In addition, the tracheal tube has a transparent outer layer that reflects light, which makes it difficult to learn, assuming that the collected dataset is in specific illumination conditions.

Recently, deep learning foundation models for image segmentation \cite{kirillov2023segment, wang2023repvitsam} have demonstrated generalizability across different scenarios and robustness against complex environments with object variance. However, the foundation model cannot be directly applied to the intubation task due to the domain gap and the deformable nature of the tracheal tube. We further combine morphological image processing techniques to overcome these challenges to track the targeted DLO. With this complete tracking pipeline, we track the tracheal tube with a rope-liked segmentation and analyze its length, curvature, and other relevant parameters.

\subsection{Imitation Learning}
Imitation learning is a method where an agent learns to accomplish a task by observing and mimicking expert demonstrations. This approach has gained significant traction recently, particularly in robotics and autonomous systems, due to its potential to teach complex behaviors.

Behavioral Cloning \cite{pomerleau1988alvinn} (BC) directly learns a mapping from states to actions using supervised learning techniques. BC suffers from compounding errors, which means small mistakes can lead to large deviations over time. Also, large amounts of data are required for complex tasks, and BC doesn't generalize well to unseen situations. 

Inverse Reinforcement Learning \cite{ng2000algorithms} (IRL) tries to infer the underlying reward function the expert is optimizing. IRL Often generalizes better than BC since it can capture the intent behind actions. The disadvantages of IRL are requiring multiple iterations of reward inference and policy optimization, and the reward function may not be uniquely determined from demonstrations.

Generative Adversarial Imitation Learning \cite{ho2016generative} (GAIL) adopts an adversarial framework to match the distribution of the learned policy to that of the expert. It can learn complex behaviors with relatively few demonstrations and performs well in high-dimensional spaces. However, it can be unstable during training due to adversarial optimization and requires careful hyperparameter tuning.

Adversarial Inverse Reinforcement Learning \cite{fu2017learning} (AIRL) combines ideas from IRL and GAIL to learn a reward function, which is robust to changes in dynamics. AIRL learns transferable reward functions and performs well with limited data. However, it needs a complex training procedure with multiple components and is sensitive to hyperparameters.

Although the robot model has shown good generalization and scene understanding capabilities, its performance in specific complex scenarios and tasks cannot meet medical safety requirements. Action chunking with transformer (ACT) \cite{zhao2023learning} learns a generative model over action sequences to address two main challenges of imitation learning: policy errors can accumulate over time, and human demonstrations may exhibit non-stationarity. This method shows the potential to accomplish complex tasks in real-world scenarios.

The main contributions of this paper are as follows:
\begin{enumerate}
    \item  An autonomous NTI system is developed, incorporating a prosthesis embedded with force sensors, allowing for safety assessment and data filtering.
    \item A novel imitation learning model, RACCT, is proposed. Compared to the ACT model, it incorporates a DLO segmentation module, an action-confidence pair sequence output structure, and a recurrent architecture, resulting in improved performance.  
\end{enumerate}
To our knowledge, this is the first work to complete autonomous NTI. What's more, the average peak contact force during autonomous NTI is only 66\% of that of the professional doctor. 


\begin{figure}[t]
    \centering

    \includegraphics[width=1\linewidth]{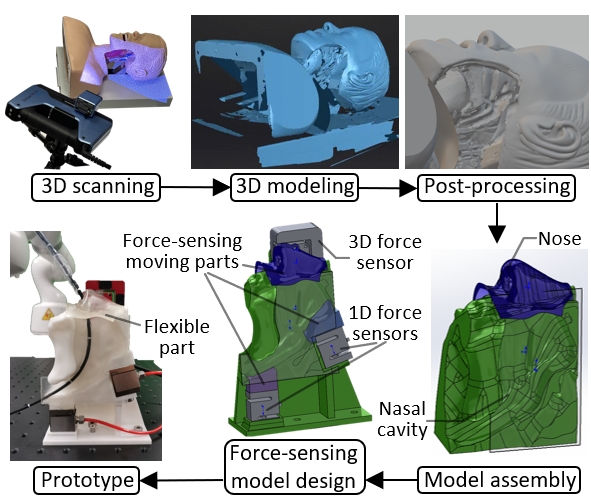}
   
    \caption{\textbf{Fabrication process of the prosthesis.}}
    \label{prosthesis}
\end{figure}

\section{System Overview}
As shown in Figure \ref{system}, the system consists of a KUKA iiwa robot and its panel, 2 cameras, a nasotracheal tube, a PC, a teleoperation controller, a prosthesis with 3 force sensors embedded in it, and the monitors of force sensors. The KUKA iiwa robot has 7 degrees of freedom, each joint is equipped with a position and torque sensor, and the end-effect is equipped with a 3D force sensor. The three force sensors are a 3D force sensor, [$F_x$, $F_y$, $F_z$], installed at the nostril, and two 1D sensors, $F_1$ and $F_2$, installed in the nasal cavity and throat. 

The fabrication process of the prosthesis is illustrated in Figure \ref{prosthesis}. It begins with 3D scanning of a commercial tracheal intubation training model (TG/J50 by Taigui Medical), which replicates human anatomical structures for clinical training scenarios, followed by 3D modeling and post-processing. Biomechanical simplification focusing on three critical luminal regions where intubation tube interactions exhibit significant force variations during insertion. Than, a force-sensing model is designed, which incorporates a 3D force sensor near the nostril, and two 1D force sensors near the sphenoid and pharynx. Multi-material 3D printing using a Stratasys J826 printer with hybrid soft materials (Shore hardness 85A, composed of Agilus30Clear with Shora-A 30, and VeroClear) for the nostril region to simulate tissue deformability, and rigid PLA for the nasal cavity areas. Finally, the assembly of the prosthesis is completed. The integrated force sensors operated as real-time force feedback is unavailable in clinical practice. Instead, force data served as data filtering and performance evaluation, rather than model inputs.

To replicate clinically challenging conditions, the phantom was partially obstructed by a white occluding box, preventing direct visual feedback of the tube’s internal state. While Camera 2 recorded the intubation process for post hoc analysis, its data was intentionally excluded from model inputs to enforce reliance on the primary endoscopic view (Camera 1), mirroring real-world surgical constraints. 

\subsection{Data Collection}
\begin{figure}[t]
  \centering
  \includegraphics[width=1\linewidth]{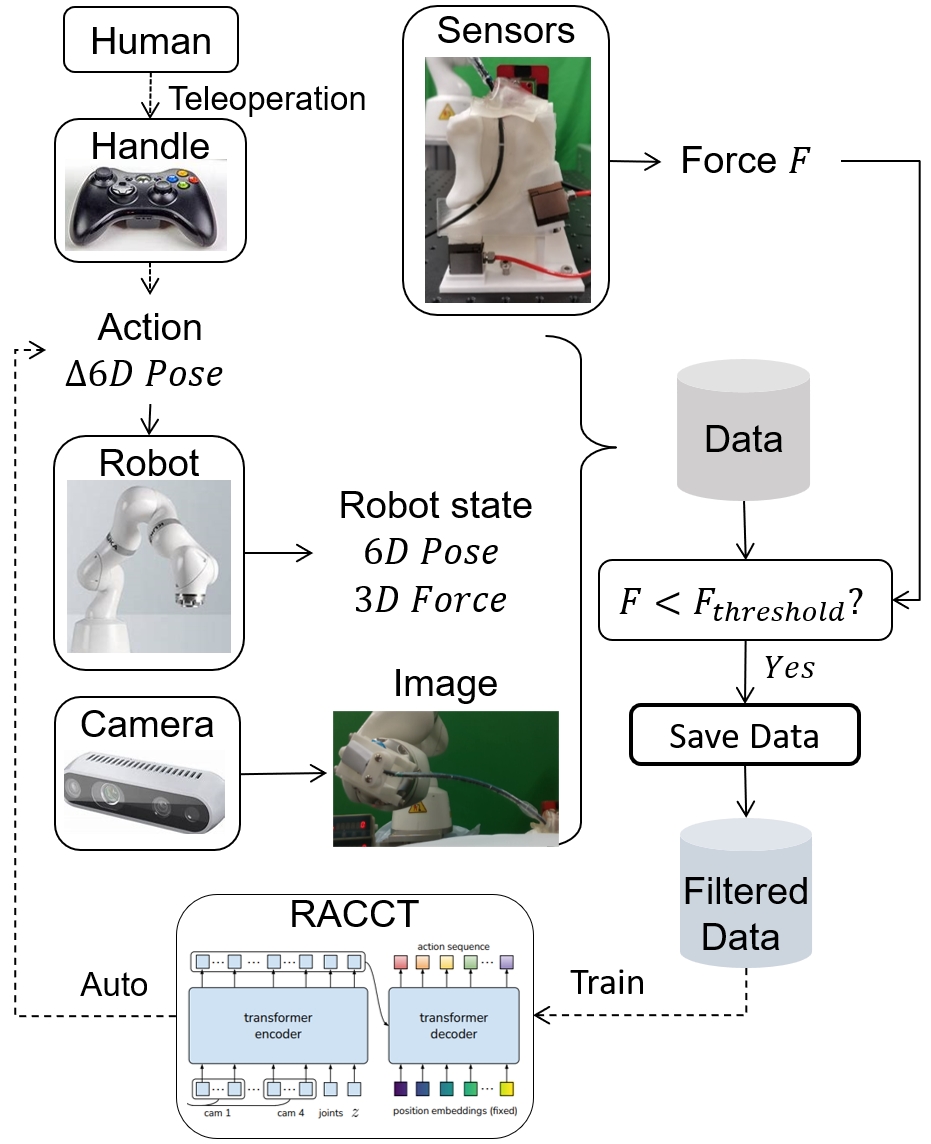}
  \caption{\textbf{System framework.} Teaching data is collected through teleoperation, and after filtering with force data, it is utilized to train the RACCT model. Finally, the model is employed to achieve autonomous intubation.}
  \label{framework}
\end{figure}

The framework of the system is shown in Figure \ref{framework}. Operators teleoperate with the robot to conduct NTI by the controller to collect data. The input of the handle is mapped to the increment of the end-effect pose of the robot, which is limited to the x-z plane, including two-dimensional translation and rotation along the y-axis, a total of three degrees of freedom. Camera 1 captures the shape of the nasotracheal tube outside the prosthesis and segments it. The contact force between the nasotracheal tube and prosthesis is collected and displayed through digital monitors. The operators can adjust the control strategy in real-time according to the contact force situation. During the process, the 6-dimensional pose and 3-dimensional force data of the robot end-effector, the image of the tube, and force data from the force sensor are recorded and saved for the following process.

\subsection{Data Filtering criteria}\label{criterial}
After each data collection episode, the data will be filtered by the following indicators: intubation time, the peak of contact force, and the impulse of contact force, that is:
\begin{equation}
I = \int \max \left(F-F_{\text {threshold }}, 0 \right) \mathrm{d} t 
\end{equation}
The impulse can reflect the degree of continuous pressure the tube exerts on the prosthesis. 
We invited a professional otolaryngology clinical doctor to perform manual intubation on the prosthesis. Based on the intubation data and the suggestion of the doctor, we established the following safety threshold: 
\begin{enumerate}
    \item  The intubation time $t<20$s.
    \item  The force peak $ F_{peak}<5 $ N.
    \item  The impulse $ ln(I)<1 $ N*s with $F_{\text {threshold }}=1.5$ N.
\end{enumerate}

Only the demonstration data with metrics below 70\% of the safety threshold will be saved and used for training. 

\begin{figure*}[t]
  \centering
  \includegraphics[width=0.80\linewidth]{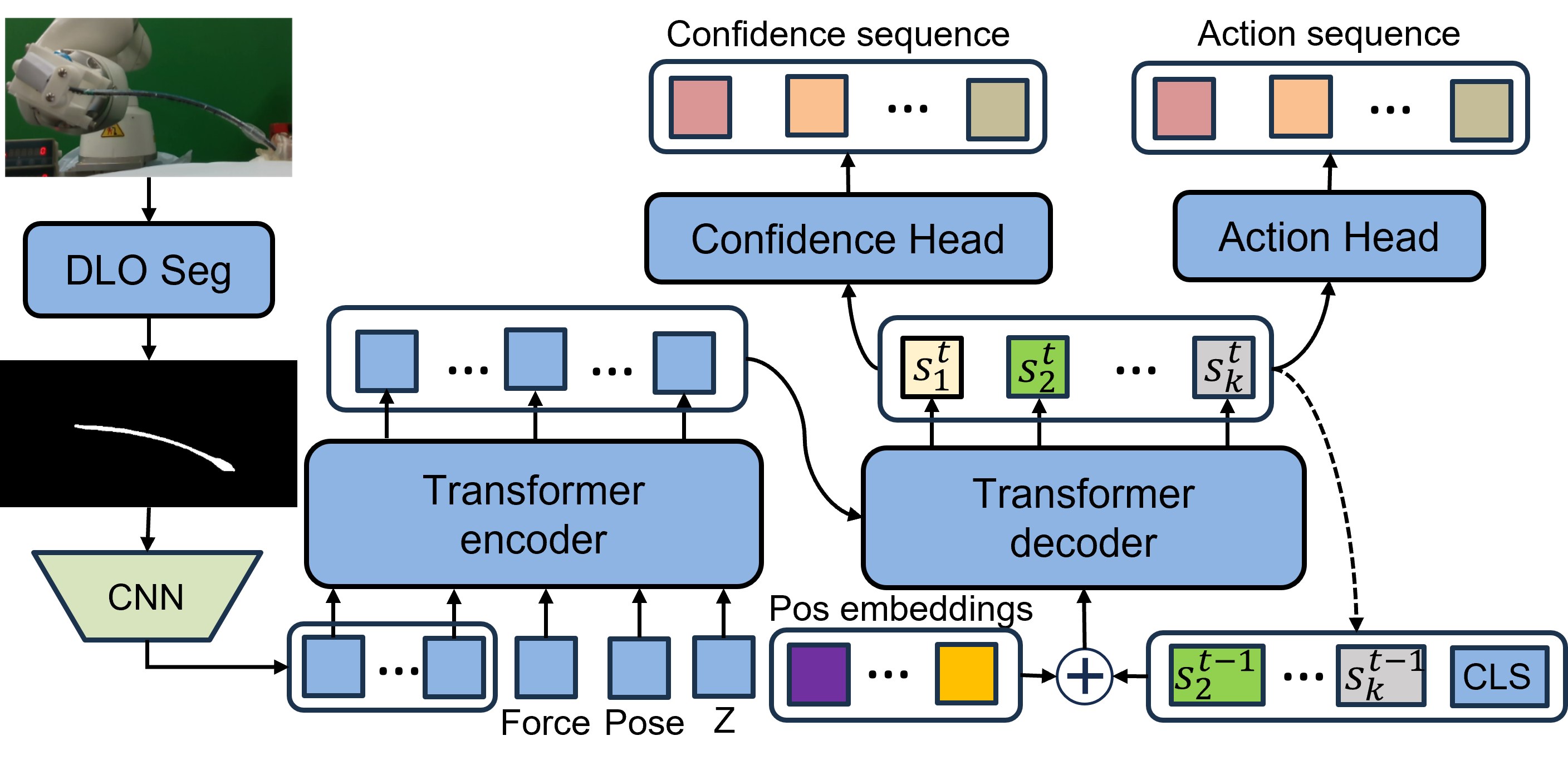}
  \caption{\textbf{Recurrent action-confidence chunking with transformer.} A tube segmentation module, an action-confidence pair sequence output structure, and a recurrent architecture decoder are added to the ACT model.}
  \label{RACCT}
\end{figure*}

\section{Methodology}
\subsection{Tube Segmentation}
To better capture the ex vivo state of the tube and eliminate redundant information from the background, we perform segmentation and tracking of the tube. We first adopt the RepViT-SAM~ \cite{wang2023repvitsam} to obtain the tube segmentation mask. After obtaining the coarse segmentation mask from the foundation model, we propose using morphological image processing techniques for accurate tube tracking. The morphological post-processing includes the following steps: (1) Extracting the connected components with 1-connectivity in the segmentation mask. (2) Remove small components with an area less than or equal to the threshold. (3) Applying a binary blob thinning operation to get the skeleton of each component. (4) Filtering the components by their width, length, number of ends, and junctions. (5) Selecting the component with the largest mean of $x$ coordinates.

With these steps, we extract the only connected component with a thin-curved skeleton of the DLO segmentation. With the skeleton of the DLO, we utilize the \textit{scipy} library for further analysis. We fit the skeleton curvature $\kappa$ with a quadratic equation of $ax^2+bx+c=y$, and further define a curvature score $s_\kappa$ as follows:
\begin{equation}  \kappa = \frac{2|a|}{\sqrt{(1+2ax+b)^3}}, \ s_\kappa = \frac{1}{1+\frac{1}{|P|}\sum_{i\in P}\kappa_i}
\end{equation}

\noindent where $P$ is the set of points lying on the skeleton curve.

\subsection{Recurrent Action-Confidence Chunking with Transformer}
The ACT model is selected as the backbone for its capability to accomplish complex tasks effectively, for example, cooking shrimp and wiping wine \cite{fu2024mobile}. However, the ability of the ACT model to perform tasks in scenarios with incomplete state observability has not been validated. In the context of NTI, a significant challenge is that the state of the tube inside the nasal cavity cannot be directly observed. To address this challenge, as illustrated in Figure \ref{RACCT}, the following improvements are made to the ACT model: an action-confidence pair sequence output structure along with a corresponding loss function is proposed, and a recurrent architecture decoder is established.

The original encoder from the ACT model has been retained. The contact forces of the prosthesis are not included as input, since this information cannot be obtained in real-world scenarios. To better estimate the contact force and the state of the tube within the nasal cavity, three-dimensional force information from the robotic arm's end-effector has been incorporated into the input.  $Z$ is the style variable of training data, which includes the distribution information and is explained in detail in  \cite{zhao2023learning}.

\subsubsection{Action-Confidence Chunking with Transformer}
In  \cite{zhao2023learning}, a temporal ensemble is proposed to overlap different action chunks with each other to improve smoothness and avoid discrete switching, while a fixed weighted scheme results in insufficient robustness of the algorithm. A significant prediction error during a particular instance can have a prolonged impact on the action outcomes for an extended period. To handle this problem, an action-confidence pair sequence output structure has been proposed. While predicting the action sequence, the corresponding confidence sequence is also outputted through a confidence head and a sigmoid function. This confidence can be regarded as an adaptive weight, enhancing the model's tolerance to errors and noise. The final actions are calculated using the following formula:
\begin{equation}
p_t = \sum_{i}(e^{-mi} C_t[i] A_t [i])/\sum_{i}(e^{-mi} C_t[i])
\end{equation}
The loss is designed as: 
\begin{equation}
Loss=\sum_{i=t}^{t+k-1}{\frac{c_i|\hat{a_i}-a_i|}{k(\epsilon+1-c_i)}}-\lambda log(\sum_{i=t}^{t+k-1}{\frac{c_i}{k}})
\end{equation}
where $p_t$ is the action that the robot will execute at time $t$. $A_t [i]$ and $C_t [i]$ are the action and confidence predicted by model for time $t$ at step $i$. $a_i$ and $c_i$ are the $i-th$ results in the current output sequence, and $\hat{a_i}$ is the ground-truth. $m$, $\epsilon$ and $\lambda$ are small positive constants, $k$ is the chunking size. 

The first term of loss is used to reduce the prediction error of the model and to limit its overconfidence. This establishes an inverse relationship between the error of the model and its level of confidence. The second term is intended to prevent the confidence from being too low. Without this constraint, both the final confidence level and the first term of the loss would converge to zero. In the second item, the reason for calculating the average first and then applying the logarithm is to ensure that the variance of the confidence is relatively larger. If the order is reversed, the confidence values across the sequence will become quite similar.

\subsubsection{Recurrent Structure}
In NTI scenarios, the state of the tube can not be fully observed. The ACT model predicts the future action sequence solely based on the current observation, which is insufficient to estimate the state of the tube. In action chunking, long-term action predictions include a wealth of temporal information. Using this information can better address the issue of incomplete observability under limited observations.  Ideally, predictions of future actions at different times should remain consistent. Such consistency can reduce contradictions between predictions at different times, leading to smoother actions. To achieve the aforementioned goals, we propose a shift-recurrent architecture. As shown in figure \ref{RACCT}, the consistency can be expressed as: $s_{k}^{t-1}=s_{k-1}^{t}$, the input to the decoder is constructed by taking the output from the previous time step, discarding the leftmost token, and appending a CLS token at the far right. Finally, position embeddings are added to form the resulting sequence. This architecture is designed to effectively integrate historical information and maintain consistency in action predictions over time, ultimately enhancing the overall performance in state estimation and action-confidence prediction.

\begin{figure}[t]
    \centering
    \includegraphics[width=1\linewidth]{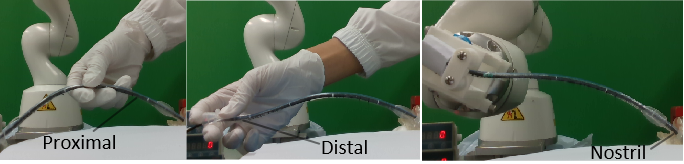}
    \caption{\textbf{Intubation methods.} From left to right, they are Manual proximal intubation, Manual distal intubation, and robotic intubation.}
    \label{Intubation methods}
\end{figure}

\section{Experiment}

\subsection{Experiment Setting}
The comparative experiments are conducted among auto-intubation methods and three groups of human participants, namely professional otolaryngology clinical doctors,  novices with no intubation experience, and experienced operators who collected the training data and have at least two hundred successful intubation experiences. 

\begin{table*}[t]
\centering
\caption{Experiment Result}
\label{tr}
\begin{tabular}{lcccccccccccc}
\hline
                                                & \multirow{2}{*}{Success Rate} & \multirow{2}{*}{Time (s)}  & \multicolumn{5}{c}{Impulse (log(N*s))}                & \multicolumn{5}{c}{Force Peak (N)} \\ \cline{4-13} 
                                                &                               &                            & $F_x$    & $F_y$    & $F_z$    & $F_1$    & $F_2$                         & $F_x$    & $F_y$    & $F_z$    & $F_1$    & $F_2$   \\ \hline
\multicolumn{1}{l|}{Novice distal intubation}   & \multicolumn{1}{c|}{95\%}     & \multicolumn{1}{c|}{5.56}  & 0.77 & 0.04 & \underline{0.00} & 0.08 & \multicolumn{1}{c|}{0.28} & 3.02  & 1.37  & 0.88 & 1.09 & 3.06 \\ \hline
\multicolumn{1}{l|}{Novice proximal intubation} & \multicolumn{1}{c|}{90\%}     & \multicolumn{1}{c|}{5.57}  & 0.31 & 0.02 & \underline{0.00} & 0.04 & \multicolumn{1}{c|}{0.26} & 2.27  & 1.25  & 0.76 & 0.86 & 2.62 \\ \hline
\multicolumn{1}{l|}{Doctor proximal intubation} & \multicolumn{1}{c|}{\underline{100\%}}    & \multicolumn{1}{c|}{\underline{5.03}}  & 0.32 & \underline{0.00} & \underline{0.00} & 0.02 & \multicolumn{1}{c|}{\underline{0.21}} & 2.75  & 1.24  & \underline{0.62} & \underline{0.40} & 2.66 \\ \hline
\multicolumn{1}{l|}{Novice teleoperation}       & \multicolumn{1}{c|}{35\%}     & \multicolumn{1}{c|}{14.20} & 0.76 & 0.03 & 0.11 & 0.48 & \multicolumn{1}{c|}{0.36} & 3.87  & 1.48  & 1.65 & 2.15 & 2.29 \\ \hline
\multicolumn{1}{l|}{Experienced teleoperation}  & \multicolumn{1}{c|}{\underline{100\%}}    & \multicolumn{1}{c|}{9.87}  & 0.30 & \underline{0.00} & 0.04 & 0.08 & \multicolumn{1}{c|}{0.33} & 2.14  & \underline{1.12}  & 1.36 & 1.97 & 2.28 \\ \hline
\multicolumn{1}{l|}{ACT}                        & \multicolumn{1}{c|}{80\%}     & \multicolumn{1}{c|}{10.24} & 0.39 & 0.12 & \underline{0.00} & 0.05 & \multicolumn{1}{c|}{0.50} & 2.41  & 1.76  & 0.93 & 1.47 & 2.36 \\ \hline
\multicolumn{1}{l|}{ACCT}                       & \multicolumn{1}{c|}{85\%}     & \multicolumn{1}{c|}{9.58}  & \underline{0.07} & 0.08 & \underline{0.00} & \underline{0.00} & \multicolumn{1}{c|}{0.41} & 1.57  & 1.62  & 1.17 & 1.02 & 2.05 \\ \hline
\multicolumn{1}{l|}{RACT}                       & \multicolumn{1}{c|}{90\%}     & \multicolumn{1}{c|}{9.85}  & 0.19 & 0.02 & \underline{0.00} & 0.01 & \multicolumn{1}{c|}{0.36} & 2.00  & 1.44  & 0.89 & 1.24 & 1.99 \\ \hline
\multicolumn{1}{l|}{\textbf{RACCT(ours)}}       & \multicolumn{1}{c|}{\underline{100\%}}    & \multicolumn{1}{c|}{9.13}  & 0.13 & 0.01 & \underline{0.00} & \underline{0.00} & \multicolumn{1}{c|}{0.31} & \underline{1.56}  & 1.51  & 1.33 & 1.09 & \underline{1.81} \\ \hline
\end{tabular}

\end{table*}

\begin{figure*} [t]
    \centering
  \includegraphics[width=0.95\linewidth]{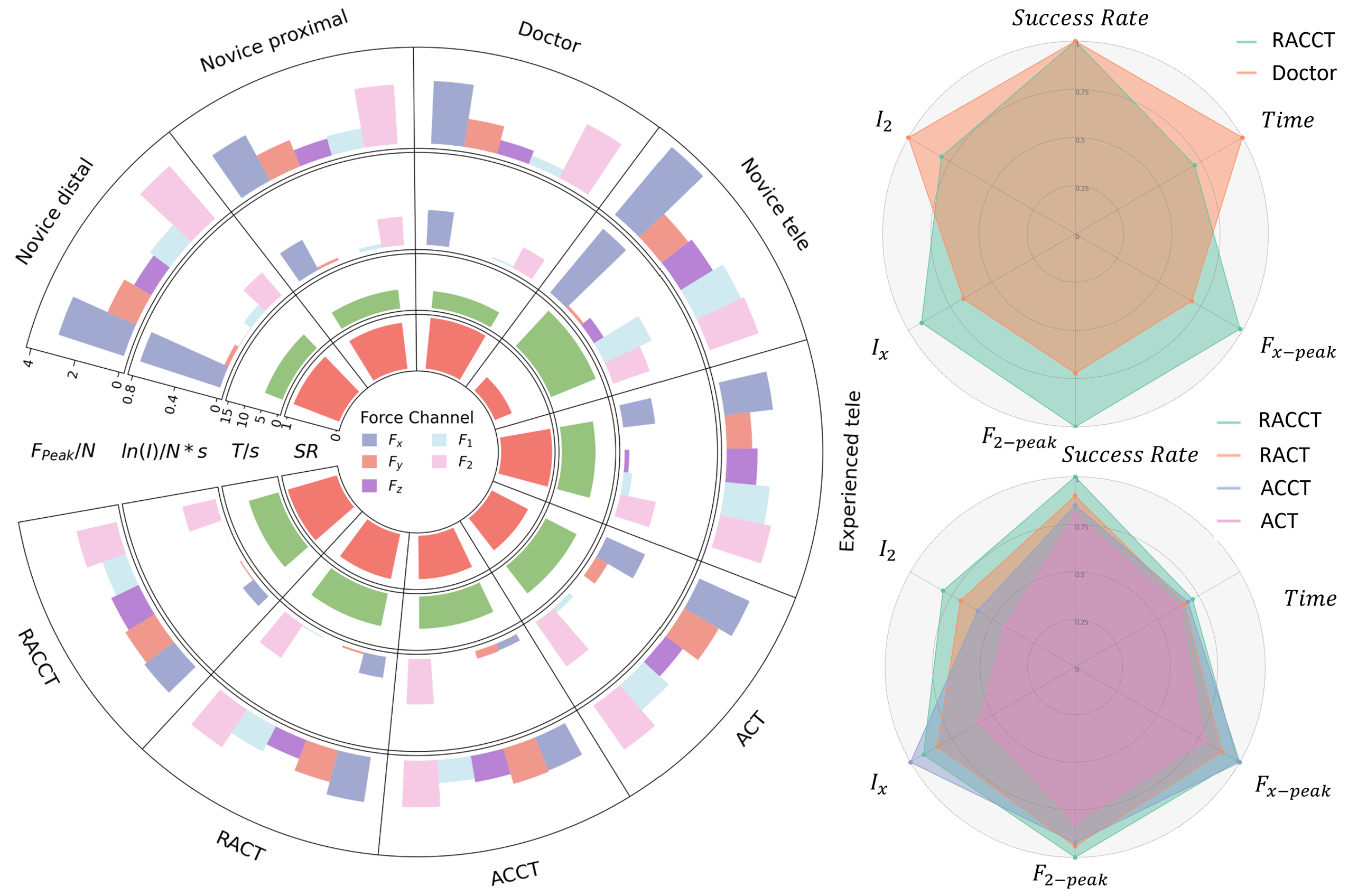}
  \caption{Experiment Result (In the left graph, the two axes represent 9 groups of comparative experiments and 4 evaluation parameters. The data in the radar chart on the right has been transposed and normalized, such that a larger area indicates better performance of the algorithm.)}
  \label{result}
\end{figure*}

As shown in Figure \ref{Intubation methods}, three different intubation methods are involved: manual proximal intubation, manual distal intubation, and robotic teleoperation intubation. Professional doctors only performed manual proximal intubation. Experienced operators and novices performed all 3 intubation methods. One professional doctor, 5 novices, and 2 experienced operators are invited to participate in the experiment. Each intubation method is repeated 20 times for each participant. 

For auto-intubation, 4 groups of ablation experiments are conducted: the original ACT model, the ACT model with action-confidence pair sequence output (ACCT), the ACT model with recurrent architecture (RACT), and the RACCT model. Each model has around 80M parameters and is trained with 50 episodes of high-quality training data for 20000 steps with $batch \, size=8,$  $chunking \, size=80$, $learning \, rate=1e^{-5}$, $m=0.95$, $\epsilon=0.2$, $\lambda=0.1$ on an RTX A6000 GPU for around 1 hour. 

For each experiment, the success rate, intubation time, peak value, and logarithmic impulse of contact force are recorded and compared. The last three metrics exclude the data from failed episodes; therefore, the success rate is relatively more important compared to the other metrics. The success criteria are outlined in the section \ref{criterial}

\subsection{Experiment Result}

All experiments were conducted on a workstation equipped with an Intel Core i9-14900K CPU, 48GB RAM, and an NVIDIA RTX 3060 GPU (12GB VRAM). RACCT achieved a real-time inference speed of 5Hz (200ms per frame), which meets the requirement for clinical applications. 

The experiment result is shown in Table \ref{tr} and Figure \ref{result}. Due to the presence of many extreme values in the failure data, only the average values derived from the successful data are presented. The left chart presents the complete data, from which it is evident that in the five force channels (direction and position of each force channel can be found in Figure \ref{system}. (b)), $F_x$ and $F_2$ are significantly greater than the other three forces, making them more likely to cause injury during the intubation process. Therefore, in the radar chart on the right, we only selected the peak values and impulses of these two channels for comparison.

Among all nine experimental groups, the teleoperation performed by novices differs significantly from the other groups; therefore, no further comparisons will be made regarding this group in the following sections. The significant differences in performance between novice and experienced operators highlight that teleoperated NTI is a task that requires extensive training and proficiency. Therefore, developing a high-quality automated intubation robot system would be highly beneficial for assisting novice doctors in mastering intubation techniques and alleviating the shortage of medical resources.

In terms of success rate, intubation performed by professional doctors, experienced operators, and RACCT achieved 100\%. This metric is crucial in the medical field as it ensures patient safety.

As for intubation time, manual intubation is much better than robotic intubation, taking only slightly more than 5 seconds. To ensure the quality of intubation, the speed of the robotic arm is set to be relatively slow. The intubation time for robotic intubation is around 10 seconds, which is still significantly below the established safety time threshold of 20 seconds. 

In terms of force peak, the intubation performed by the doctor has the lowest $F_z$ and  $F_1$ peak. The experienced teleoperation has the lowest $F_y$ peak. RACCT intubation performed best for $F_x$ and $F_2$ peak, the 2 force channels with the highest peak, which pose the greatest risk of harm to patients. So it is crucial to reduce the stress on these 2 channels.

As for the impulse of contact force, RACCT exhibits the lowest force impulse in channels $F_z$, and $F_1$, while in channel $F_x$, it is slightly higher than ACCT, ranking second. The intubation performed by the doctor shows a better impulse of $F_y$ and $F_2$, even though its peak of $F_2$ is larger than that of RACCT. Considering the differences in intubation time, the ability to maintain such a low level of impulse indicates that the duration of significant contact forces is relatively short. 

From the radar chart in the upper right, we can see that the intubation performance of RACCT is comparable to that of the doctor. RACCT has a disadvantage in terms of time, but it performs better than the doctor in terms of force peak and impulse of $F_x$. Comparing the peak values of all force channels, the peak force of the doctor is in channel $F_x$, 2.75 N, while the peak of RACCT is in channel $F_2$, 1.81 N, which is only 66\% of the peak of the doctor. In terms of safety, the RACCT-based automatic NTI platform we have developed is approaching the level of performance exhibited by the doctor. 

The radar chart below further demonstrates the significant improvements achieved. The data shows that both the confidence architecture and the recurrent architecture significantly optimize the ACT model, with the recurrent architecture providing a greater improvement than the confidence architecture. With the combination of the two architectures, RACCT outperforms ACT in various aspects, thereby confirming the effectiveness of our proposed method.

\section{Conclusion}
This paper presents a low-contact force autonomous nasotracheal intubation system. Thanks to the contact force quality filtering system and the newly proposed imitation learning model, RACCT, the intubation performance of this system is approaching that of the doctor in terms of contact force. Besides, the improved RACCT model outperforms the ACT model in all aspects. For future work, We will conduct animal experiments and try to use the system in real human bodies. In addition, we will try to generalize the method to the insertion tasks of tubular objects in other scenarios, for example, the intervention of gastrointestinal endoscopy.



\bibliography{ref}

\begin{thebibliography}{10}
\providecommand{\url}[1]{#1}
\csname url@samestyle\endcsname
\providecommand{\newblock}{\relax}
\providecommand{\bibinfo}[2]{#2}
\providecommand{\BIBentrySTDinterwordspacing}{\spaceskip=0pt\relax}
\providecommand{\BIBentryALTinterwordstretchfactor}{4}
\providecommand{\BIBentryALTinterwordspacing}{\spaceskip=\fontdimen2\font plus
\BIBentryALTinterwordstretchfactor\fontdimen3\font minus \fontdimen4\font\relax}
\providecommand{\BIBforeignlanguage}[2]{{%
\expandafter\ifx\csname l@#1\endcsname\relax
\typeout{** WARNING: IEEEtran.bst: No hyphenation pattern has been}%
\typeout{** loaded for the language `#1'. Using the pattern for}%
\typeout{** the default language instead.}%
\else
\language=\csname l@#1\endcsname
\fi
#2}}
\providecommand{\BIBdecl}{\relax}
\BIBdecl

\bibitem{torrego2020bronchoscopy}
A.~Torrego, V.~Pajares, C.~Fern{\'a}ndez-Arias, P.~Vera, and J.~Mancebo, ``Bronchoscopy in patients with covid-19 with invasive mechanical ventilation: A single-center experience,'' \emph{Am. J. Respir. Crit. Care Med.}, vol. 202, no.~2, pp. 284--287, 2020.

\bibitem{tobin2020basing}
M.~J. Tobin, ``Basing respiratory management of covid-19 on physiological principles,'' pp. 1319--1320, 2020.

\bibitem{deng2024assisted}
Z.~Deng, S.~Zhang, Y.~Guo, H.~Jiang, X.~Zheng, and B.~He, ``Assisted teleoperation control of robotic endoscope with visual feedback for nasotracheal intubation,'' \emph{Robotics and Autonomous Systems}, vol. 172, p. 104586, 2024.

\bibitem{murakami2023therapeutic}
N.~Murakami, R.~Hayden, T.~Hills, H.~Al-Samkari, J.~Casey, L.~D. Sorbo, P.~R. Lawler, M.~E. Sise, and D.~E. Leaf, ``Therapeutic advances in covid-19,'' \emph{Nat. Rev. Nephrol.}, vol.~19, no.~1, pp. 38--52, 2023.

\bibitem{licker2007perioperative}
M.~Licker, A.~Schweizer, C.~Ellenberger, J.-M. Tschopp, J.~Diaper, and F.~Clergue, ``Perioperative medical management of patients with copd,'' \emph{International Journal of Chronic Obstructive Pulmonary Disease}, vol.~2, no.~4, pp. 493--515, 2007.

\bibitem{christian2020use}
C.~E. Christian, N.~E. Thompson, and M.~K. Wakeham, ``Use and outcomes of nasotracheal intubation among patients requiring mechanical ventilation across us picus,'' \emph{Pediatric Critical Care Medicine}, vol.~21, no.~7, pp. 620--624, 2020.

\bibitem{chauhan2016nasal}
V.~Chauhan and G.~Acharya, ``Nasal intubation: A comprehensive review,'' \emph{Indian J. Crit. Care Med.}, vol.~20, no.~11, p. 662, 2016.

\bibitem{nan2023characteristics}
R.~Nan, Y.~Su, J.~Pei, H.~Chen, L.~He, X.~Dou, and S.~Nan, ``Characteristics and risk factors of nasal mucosal pressure injury in icus,'' \emph{J. Clin. Nurs.}, vol.~32, no. 1-2, pp. 346--356, 2023.

\bibitem{sun2010cardiovascular}
Y.~Sun, J.-X. Liu, H.~Jiang, Y.-S. Zhu, H.~Xu, and Y.~Huang, ``Cardiovascular responses and airway complications following awake nasal intubation with blind intubation device and fibreoptic bronchoscope: a randomized controlled study,'' \emph{European Journal of Anaesthesiology (EJA)}, vol.~27, no.~5, pp. 461--467, 2010.

\bibitem{gasmi2022improving}
A.~G. Benahmed, A.~Gasmi, W.~Anzar, M.~Arshad, and G.~Bj{\o}rklund, ``Improving safety in dental practices during the covid-19 pandemic,'' \emph{Health and technology}, vol.~12, no.~1, pp. 205--214, 2022.

\bibitem{izzetti2020covid}
R.~Izzetti, M.~Nisi, G.~Gabriele, and F.~Graziani, ``Covid-19 transmission in dental practice: brief review of preventive measures in italy,'' \emph{J. Dent. Res.}, vol.~99, no.~9, pp. 1030--1038, 2020.

\bibitem{dupont2021decade}
P.~E. Dupont, B.~J. Nelson, M.~Goldfarb, B.~Hannaford, A.~Menciassi, M.~K. O’Malley, N.~Simaan, P.~Valdastri, and G.~Yang, ``A decade retrospective of medical robotics research from 2010 to 2020,'' \emph{Science Robotics}, vol.~6, no.~60, p. eabi8017, 2021.

\bibitem{ng2024navigation}
C.~Ng, H.~Gao, T.-A. Ren, J.~Lai, and H.~Ren, ``Navigation of tendon-driven flexible robotic endoscope through deep reinforcement learning,'' in \emph{2024 IEEE Int. Conf. Adv. Robot. Soc. Impacts (ARSO)}.\hskip 1em plus 0.5em minus 0.4em\relax IEEE, 2024, pp. pp. 134--139.

\bibitem{hemmerling2012first}
T.~M. Hemmerling, R.~Taddei, M.~Wehbe, C.~Zaouter, S.~Cyr, and J.~Morse, ``First robotic tracheal intubations in humans using the kepler intubation system,'' \emph{Br. J. Anaesth.}, vol. 108, no.~6, pp. 1011--1016, 2012.

\bibitem{wang2018original}
X.~Wang, Y.~Tao, X.~Tao, J.~Chen, Y.~Jin, Z.~Shan, J.~Tan, Q.~Cao, and T.~Pan, ``An original design of remote robot-assisted intubation system,'' \emph{Sci. Rep.}, vol.~8, no.~1, p. 13403, 2018.

\bibitem{liang2020pneumatic}
Z.~Liang, H.~Miao, Z.~Guo, X.~Zhu, X.~Wang, Q.~Cao, X.~Tao, and T.~Pan, ``Pneumatic actuator based tracheal intubation system,'' in \emph{Proceedings of 2020 3rd World Conf. Mech. Eng. Intell. Manuf. (WCMEIM)}, 2020, pp. pp. 810--814.

\bibitem{ponraj2022chip}
G.~Ponraj, C.~J. Cai, and H.~Ren, ``Chip-less real-time wireless sensing of endotracheal intubation tubes by printing and mounting conformable antenna tag,'' \emph{IEEE Robotics and Automation Letters}, vol.~7, no.~2, pp. 2369--2376, 2022.

\bibitem{lai2023sim}
J.~Lai, T.-A. Ren, W.~Yue, S.~Shijian, J.~Y.~K. Chan, and H.~Ren, ``Sim-to-real transfer of soft robotic navigation strategies that learns from the virtual eye-in-hand vision,'' \emph{IEEE Transactions on Industrial Informatics}, 2023.

\bibitem{wang2023domain}
G.~Wang, T.-A. Ren, J.~Lai, L.~Bai, and H.~Ren, ``Domain adaptive sim-to-real segmentation of oropharyngeal organs,'' \emph{Medical \& Biological Engineering \& Computing}, vol.~61, no.~10, pp. 2745--2755, 2023.

\bibitem{deng2023safety}
Z.~Deng, P.~Jiang, Y.~Guo, S.~Zhang, Y.~Hu, X.~Zheng, and B.~He, ``Safety-aware robotic steering of a flexible endoscope for nasotracheal intubation,'' \emph{Biomedical Signal Processing and Control}, vol.~82, p. 104504, 2023.

\bibitem{deng2023automatic}
Z.~Deng, X.~Wei, X.~Zheng, and B.~He, ``Automatic endoscopic navigation based on attention-based network for nasotracheal intubation,'' \emph{Biomedical Signal Processing and Control}, vol.~86, p. 105035, 2023.

\bibitem{hao2025variable}
R.~Hao, J.~Lai, W.~Zhong, D.~Xie, Y.~Tian, T.~Zhang, Y.~Zhang, C.~P.~L. Chan, J.~Y.~K. Chan, and H.~Ren, ``Variable-stiffness nasotracheal intubation robot with passive buffering: a modular platform in mannequin studies,'' in \emph{2025 IEEE International Conference on Robotics and Automation (ICRA)}.\hskip 1em plus 0.5em minus 0.4em\relax IEEE, 2025.

\bibitem{zanella2021auto}
R.~Zanella, A.~Caporali, K.~Tadaka, D.~D. Gregorio, and G.~Palli, ``Auto-generated wires dataset for semantic segmentation with domain-independence,'' in \emph{2021 International conference on computer, control and robotics (ICCCR)}.\hskip 1em plus 0.5em minus 0.4em\relax IEEE, 2021, pp. 292--298.

\bibitem{thananjeyan2022all}
B.~Thananjeyan, J.~Kerr, H.~Huang, J.~E. Gonzalez, and K.~Goldberg, ``All you need is luv: Unsupervised collection of labeled images using uv-fluorescent markings,'' in \emph{2022 IEEE/RSJ Int. Conf. Intelligent Robots and Systems (IROS)}.\hskip 1em plus 0.5em minus 0.4em\relax IEEE, 2022, pp. 3241--3248.

\bibitem{caporali2022fastdlo}
A.~Caporali, K.~Galassi, R.~Zanella, and G.~Palli, ``Fastdlo: Fast deformable linear objects instance segmentation,'' \emph{IEEE Robotics and Automation Letters}, vol.~7, no.~4, pp. 9075--9082, 2022.

\bibitem{kirillov2023segment}
A.~Kirillov, E.~Mintun, N.~Ravi, H.~Mao, L.~Rolland, C.~Gustafson, T.~Xiao, S.~Berg, A.~C. Lo, W.~Yen \emph{et~al.}, ``Segment anything,'' in \emph{Proceedings of the IEEE/CVF Int. Conf. Comput. Vis.}, 2023, pp. 4015--4026.

\bibitem{wang2023repvitsam}
A.~Wang, H.~Chen, Z.~Lin, J.~Han, and G.~Ding, ``Repvit-sam: Towards real-time segmenting anything,'' 2023.

\bibitem{pomerleau1988alvinn}
D.~A. Pomerleau, ``Alvinn: An autonomous land vehicle in a neural network,'' \emph{Adv. Neural Inf. Process. Syst.}, vol.~1, 1988.

\bibitem{ng2000algorithms}
A.~Y. Ng, S.~Russell \emph{et~al.}, ``Algorithms for inverse reinforcement learning.'' in \emph{ICML}, vol.~1, no.~2, 2000, p.~2.

\bibitem{ho2016generative}
J.~Ho and S.~Ermon, ``Generative adversarial imitation learning,'' \emph{Adv. Neural Inf. Process. Syst.}, vol.~29, 2016.

\bibitem{fu2017learning}
J.~Fu, K.~Luo, and S.~Levine, ``Learning robust rewards with adversarial inverse reinforcement learning,'' \emph{arXiv preprint arXiv:1710.11248}, 2017.

\bibitem{zhao2023learning}
T.~Z. Zhao, V.~Kumar, S.~Levine, and C.~Finn, ``Learning fine-grained bimanual manipulation with low-cost hardware,'' \emph{arXiv preprint arXiv:1710.11248}, 2023.

\bibitem{fu2024mobile}
Z.~Fu, T.~Z. Zhao, and C.~Finn, ``Mobile aloha: Learning bimanual mobile manipulation with low-cost whole-body teleoperation,'' \emph{arXiv preprint arXiv:2401.02117}, 2024.

\end{thebibliography}


\end{document}